\documentclass[runningheads]{llncs}

\makeatletter
\renewcommand\subsubsection{\@startsection{subsubsection}{3}{\z@}%
                       {-18\p@ \@plus -4\p@ \@minus -4\p@}%
                       {0.5em \@plus 0.22em \@minus 0.1em}%
                       {\normalfont\normalsize\bfseries\boldmath}}
\makeatother
\usepackage{amssymb}
\usepackage{pifont}
\usepackage{lipsum}
\usepackage{booktabs}
\usepackage{graphicx}
\usepackage{array} 
\usepackage{tabularray}
\usepackage{rotating}
\usepackage{lscape} 
\usepackage{hyperref}
\usepackage{float}
\usepackage{placeins}
\usepackage{multirow} 
\usepackage{adjustbox} 
\usepackage{amsmath} 
\usepackage{amsfonts} 
\usepackage[T1]{fontenc}
\usepackage{graphicx}
\begin{document}
\title{Attend, Distill, Detect: Attention-aware Entropy Distillation for Anomaly Detection}
\titlerunning{Attend, Distill, Detect}
%
\author{Sushovan Jena \inst{1} \and
Vishwas Saini\inst{1} \and
Ujjwal Shaw\inst{1} \and Pavitra Jain\inst{1} \and Abhay Singh Raihal\inst{1} \and Anoushka Banerjee\inst{2} \and Sharad Joshi\inst{2} \and Ananth Ganesh\inst{2} \and Arnav Bhavsar\inst{1}}
\authorrunning{Sushovan et al.}
%
\institute{School of Computing and Electrical Eng., Indian Institute of Technology Mandi, 175005 Himachal Pradesh , India \and
R\&D Center, Hitachi India Pvt. Ltd., 560055 Bengaluru, India \\
\email{\{sushovanjena, vis.saini10, ujjwalshaw2002, pavitrajain29112002, abhaysinghraihal1\}@gmail.com, \{anoushka.banerjee, sharad.joshi, ananth.ganesh\}@hitachi.co.in, arnav@iitmandi.ac.in}\\
}
 \maketitle             
\vspace{-10pt}
\begin{abstract}

Unsupervised anomaly detection encompasses diverse applications in industrial settings where a high-throughput and precision is imperative. Early works were centered around one-class-one-model paradigm, which poses significant challenges in large-scale production environments. Knowledge-distillation based multi-class anomaly detection promises a low latency with a reasonably good performance but with a significant drop as compared to one-class version. We propose a \textit{DCAM} (Distributed Convolutional Attention Module) which improves the distillation process between teacher and student networks when there is a high variance among multiple classes or objects. Integrated multi-scale feature matching strategy to utilise a mixture of multi-level knowledge from the feature pyramid of the two networks, intuitively helping in detecting anomalies of varying sizes which is also an inherent problem in the multi-class scenario. Briefly, our \textit{DCAM} module consists of Convolutional Attention blocks distributed across the feature maps of the student network, which essentially learns to masks the irrelevant information during student learning alleviating the "cross-class interference" problem. This process is accompanied by minimizing the relative entropy using KL-Divergence in Spatial dimension and a Channel-wise Cosine Similarity between the same feature maps of teacher and student. The losses enables to achieve scale-invariance and capture non-linear relationships. We also highlight that the DCAM module would only be used during training and not during inference as we only need the learned feature maps and losses for anomaly scoring and hence, gaining a performance gain of \textbf{\textit{3.92\%}} than the multi-class baseline with a preserved latency.

\end{abstract}


%
%
%
\section{Introduction}

 Anomaly Detection is a highly researched field in computer vision and deep learning which has applications in defect detection \cite{IUDS,DLBDD}, visual inspection, product quality control, medical imaging, etc. That necessitates the focus on the trade-off between precision and latency constraints in low-resource environments. Anomalies or outliers are essentially open-set instances whose patterns deviate from the modeled data \cite{DLADACM,ADACM}. Earlier works focused on defect detection \cite{MDDM} involved both traditional approaches and modern deep networks and that was eventually followed by one-class methods \cite{12,13,14,15,16} where separate models are trained for a specific category of object or textures. All of these methods are trained on the normal (or non-anomalous) samples of the respective categories and detect anomalies in the same category. This inherently poses a scalability and adaptability limit where the model count escalates proportionally with the class count. This model-per-class paradigm is also least expected to perform well where there is large intra-class variation (i.e. when a class/category has more variation in objects). As a consequence, multi-class anomaly detection methods have emerged very recently where a unified model \cite{17,18} is able to serve all the classes but the latency aspects of those models were not discussed.

 In addition to generalisation across classes, we intended to emphasize on the real-time latency of such algorithms to be deployed in industrial systems, hence we explored specifically Knowledge-Distillation (KD) based anomaly detection methods \cite{19,20,STPFM}. Knowledge-Distillation as introduced in \cite{hinton2015distilling}, is a way to transfer the generalization ability of a Teacher model to a Student model on the same training set or a different dataset using the teacher's learned parameter values, logits or class probabilities. Although KD was initially used for model compression to reduce latency or complexities of models, it is also leveraged for transferring knowledge from a network trained on a large corpus of data (for example ImageNet \cite{imagenet}) to an application specific model (for MVTec AD \cite{mvtec_dataset}). Along a similar line of thought, in the case of anomaly detection, an important consideration in KD is used to bring the teacher and student embeddings closer in the feature or embedding space for the normal or good images, so that during inference when an anomalous image is passed to the teacher and student, their embeddings would differ by a good enough margin as only normal images were used during training. Then such a framework would be better suitable for unsupervised scenarios, coherently utilising the advantage of distillation in performance gain.

 Our proposed approach is a combination of Spatial and Channel-wise Attention blocks distributed across different scales of feature maps for distilling the intermediate feature information between teacher and student solving the cross-class interference that arises when dealing with multiple classes. We use cosine distance and KL divergence as loss functions for attention-aware feature matching in the student-teacher framework. Cosine distance enhances model generalizability and feature vector similarity by bringing feature vectors closer by targeting the angular distance between teacher and student features, while KL divergence captures the relative entropy and non-linear relationships between distributions of student and teacher feature maps improving feature replication between student and teacher networks. 

 Our major contributions include :-
 (i) DCAM (Distributed Convolutional Attention Module) which consists of Spatial and Channel-wise Attention which can be seamlessly integrated into a Knowledge-Distillation framework for attention-aware distillation during training, while not disturbing the inference latency.
 (ii) Analysis of KL-Divergence loss both along channel and spatial dimension for multi-scale feature distillation and its latency.
 (iii) Analysis of Cosine distance loss both along channel and spatial dimension for multi-scale feature distillation and its latency.
 (iv) Comparison of Mean-Squared Error and Cosine Distance as a metric for anomaly scoring and their latency.
 (v) The best combination of the attention modules together with the appropriate loss fucntions resulted in a 3.92\% boost in performance with a preserved latency. 


\section{Related Work}

Multi-class anomaly detection has become a crucial research area due to its real-world applications in various domains. Traditional one-class anomaly detection methods require separate models for each class \cite{12,13,14,15,16}. This approach becomes impractical for scenarios with many classes due to scalability issues and a rapidly increasing model count \cite{14}.
Distilling knowledge between two networks are experimented with various perturbations.
Bergmann et. al \cite{19} used the output logits (final layer embeddings) of the teacher as targets for the student but followed a patch-based approach which is a time-consuming strategy during inference.

Wang et. al \cite{STPFM} leveraged the intermediate feature matching strategy between teacher and student which resulted in significant gains with a reasonably low latency. Deng et. al \cite{20} introduced reverse distillation strategy where there is a teacher encoder and student decoder along with reconstructing their intermediate feature maps, which has a higher performance than the previous but again with more latency. Among the discussed methods, considering the lower latency and decent performance of the feature matching matching strategy \cite{20}, we tried to improvise upon the same for the multi-class case.

Although recently there have been good works in multi-class anomaly detection, their latency and memory-heavy architectures were not compared to their one-class counterparts, rather only the segmentation results are used for comparison.
Recent advancements have focused on multi-class anomaly detection, where a single model can handle multiple classes. You et al. \cite{17} introduced "UniAD", a transformer-based feature reconstruction model that effectively addressed this challenge. However, the inherent computational complexity and large number of parameters associated with transformers limit their practicality for resource-constrained environments \cite{17}. Additionally, Zhao et al. \cite{18} proposed "Omnial," a unified CNN framework that demonstrated promising results for multi-class anomaly detection but it involved anomaly synthesis unlike our focus on solving the problem in a completely unsupervised way for which MVTec AD is mostly designed. Lately, a work by Deng and Li \cite{deng2024structural}, showed very good improvements using the existing feature matching strategy method \cite{20} as backbone, but it consisted of four different losses and a CRAM (Central Residual Aggregation Module) during training and utilise the intra-affinity error of the teacher and student features followed by pairwise-similarity difference map for anomaly scoring, where the affinity matrix is an outer dot-product of the feature map of a layer with itself. This involves expensive element-wise multiplications of the high-dimensional feature maps which contributes to latency in a low-resource setting and adds to the difficulty of implementation.

Our proposed approach addresses the limitations of existing methods by potentially incorporating: Improvements in network architecture and loss functions for better performance, Spatial and channel-wise attention blocks to address cross-class interference during multi-class distillation, Maintaining low latency during inference despite the improvements. By leveraging these advancements, our work aims to contribute to the field of multi-class anomaly detection using knowledge distillation, offering a balance between performance and efficiency for real-world applications.

\section{Methodology}
\vspace{-20pt}
\FloatBarrier
\begin{figure}[htbp]
  \centering
  \includegraphics[width=1.05\textwidth]{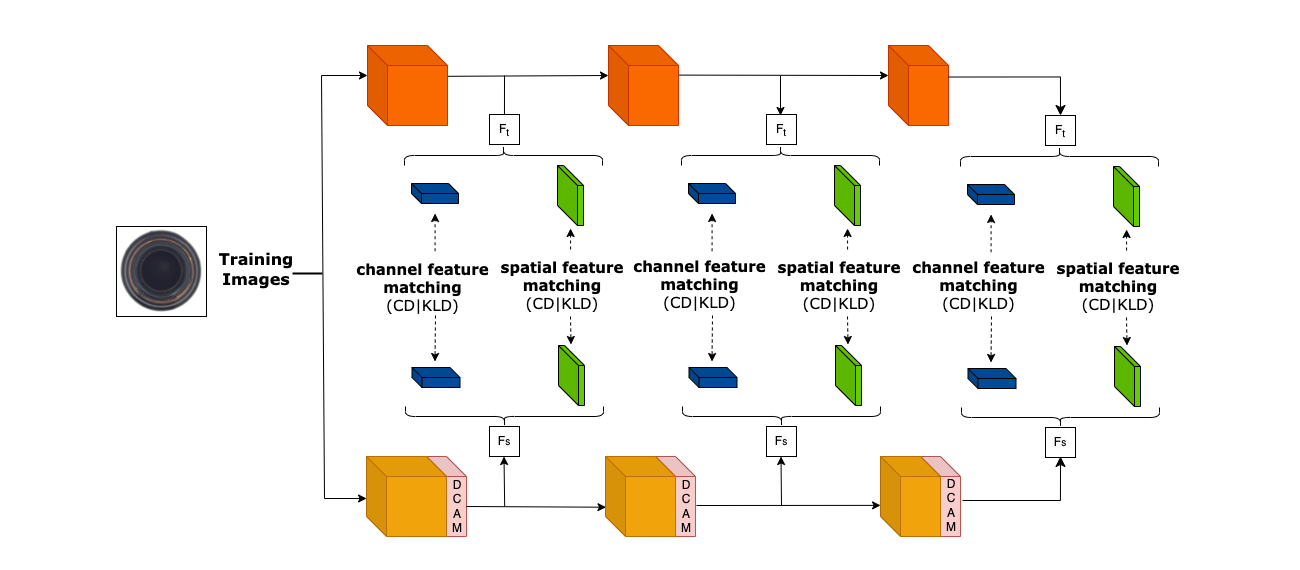}
  \caption{Overview of our teacher-student framework (Training phase). The orange and yellow blocks represent the $2^{nd}$, $3^{rd}$, and $4^{th}$ convolutional blocks of the teacher and student network respectively. During the training phase, the feature map of the student network passes through DCAM for feature refinement, followed by channel and spatial feature matching with the corresponding teacher feature maps (using cosine distance and KL divergence). }
  \label{fig:image_label}
\end{figure}
\FloatBarrier
\vspace{-10pt}
Our overall method for improving multi-class anomaly detection targets the improvisation of feature reconstruction or matching during the knowledge distillation process. We designed a Distributed Convolutional Attention Module (DCAM). DCAM distributes the attention to multiple scales of the feature pyramid both in spatial and channel dimensions of the Student network, so that instead of learning all the features, the student network learns only the vital information as the objects or classes have a high variance in the multi-class case.

In a typical student-teacher framework, a pre-trained teacher network guides the student network during its training process. The student network targets the output of the teacher network using a predefined loss metric, which was Mean Squared Error (MSE) in the case of STFPM \cite{STPFM}. Directly computing Mean Squared Error (MSE) between the student and teacher feature embeddings, fails to highlight the distinct importance of spatial and channel features, resulting in a vague understanding of the underlying data distribution. Our approach integrates spatial and channel attention mechanisms, along with different loss functions for measuring spatial and channel feature similarity between teacher and student networks.
By utilizing channel and spatial attention mechanisms, the student network learns the importance of each channel's information and spatial details at each pixel location, enhancing its ability to identify critical regions in intermediate feature maps. The DCAM module aids in mitigating cross-class interference across the dataset's 15 classes, allowing focused attention on relevant parts of the student feature maps before feature matching. We utilise these refined feature maps for knowledge distillation. 

While MSE is generally used in knowledge distillation, KL-Divergence can perform better than MSE due to its ability to capture relative entropy and non-linear relationships between distributions of student and teacher feature maps. We use the findings of prior studies that KL-Divergence is quite intuitive and effective in matching the probabilistic score between student and teacher feature maps \cite{hinton2015distilling}. Furthermore, we use cosine distance to measure the similarity between feature vectors, as it identifies the directional similarity using angular information in feature space, captures the correlation structure, and facilitates the transfer of rich knowledge from the teacher to the student network \cite{xu2020knowledge}. In the inference phase, we create the anomaly map by combining upsampled loss maps from individual blocks, computed using cosine distance between teacher and student feature maps. A detailed explanation of each method is given in the subsequent section.
\subsection{Distributed Convolutional Attention Module (DCAM)}

 Distributed Convolutional Attention Module (DCAM) proposes two components, channel attention module and spatial attention module. These attention modules intend to compute complementary attention scores essentially, to learn both the "what" and "where" aspects of the feature maps that the student network should learn. Our DCAM is inspired from the CBAM approach \cite{woo2018cbam}. The layers of a convolutional neural network consist of different channels that depict a unique or similar feature representation in terms of colour variations, texture details, edges, and so on. By utilizing the channel attention mechanism, the student network learns the channel mask that represents the importance of each channel's information. Similarly, in the spatial attention mechanism, the student network learns the spatial mask that represents the importance of the spatial information at each pixel location. It increases the ability of the student network to identify the important regions in the intermediate feature maps which has to be distilled from the teacher. Due to the multitude of data across 15 classes which gives rise to cross-class interference, our DCAM module enables better focus only on the relevant parts of the student feature maps before the feature matching step. We utilise these refined feature maps for knowledge distillation. Note that we incorporate DCAM for feature refinement only during the training process and not during the test phase, which results in minimal effect on latency of the model. 
\hfill \break
\FloatBarrier
\begin{figure}[htbp]
  \centering
  \includegraphics[width=1.0\textwidth]{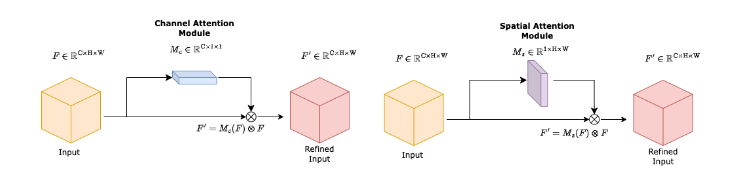}
  \caption{Overview of the Channel and Spatial attention module. F' is the refined feature map obtained after each attention block.}
  \label{fig:image_label}
\end{figure}
\FloatBarrier
\vspace{-20pt}
 Channel attention module enables the student network to prioritise informative channels by assigning different importance to each channel, through this the student network learns the fact that not all channels contribute equally to the knowledge distillation process. \hfill \vspace{4.5pt}

 Given an input feature map \( F \) (\( R^{C \times H \times W} \)), the channel attention module performs max-pooling and average-pooling across the spatial dimension. It then passes these pooled features through a shared MLP which outputs 2 separate vectors, for max-pooling and average pooling respectively. Thereafter, it aggregates the obtained vectors and passes them through a sigmoid non-linear function to produce the final 1-D channel attention map \( M_c \). This attention map indicates which channels are of more importance to the student network.\hfill \break

\[ M_c(F) = \sigma(W_1(W_0(F_{c_{\text{avg}}})) + W_1(W_0(F_{c_{\text{max}}}))) \]
 The spatial attention module enhances the student network's learning process by prioritizing informative regions within the spatial dimension and focusing on more important pixel locations. Similar to channel attention, which identifies important channels, spatial attention identifies where the crucial information resides within each channel, capturing non-local dependencies across the feature maps. This filtering of information enhances learning during the distillation process. 

 Given an input feature map  \( F \) (\( R^{C \times H \times W} \)), spatial attention module performs max-pooling and average-pooling in the channel dimension and concatenate them together. Then it convolves them with a \( 7 \times 7 \) kernel which is then passed through a sigmoid non-linear function to produce the final 2D spatial attention map \( M_s \).\hfill \break 

\[ M_s(F) = \sigma(f_{7 \times 7}([F_{s_{\text{avg}}}; F_{s_{\text{max}}}])) \]

\subsection{Cosine Distance (CD)}
We utilize cosine similarity distance to match the refined feature map of the student with the teacher's feature map, both in the spatial and channel dimensions. In prior studies, cosine distance has been effective in knowledge distillation, leading to improved performance in various applications. Cosine similarity distance is scale-invariant and captures the direction of the two feature vectors, making it an efficient loss metric for feature matching under our student-teacher framework. 

In the channel dimension, cosine distance captures the angular distance of teacher and the student features at each pixel location. Likewise, in the spatial dimension, the student network aligns the channel-wise spatial information in the angular feature space. Cosine similarity has been shown to be an effective metric when the dimensionality of data is high \cite{article} as it normalises the magnitude of the feature vectors and tries to minimize their angular distance. In our case, the intermediate feature maps have very high dimensionality consisting of 64, 128 and 256 channels respectively in the 3 layers considered for feature matching between student and teacher. This ensures the elimination of redundant and irrelevant features.


Let \( T_{feat}^{k} \) and \( S_{feat}^{k} \) represent the \(k^{th}\) feature maps of the teacher and student models, respectively. The feature maps are represented as tensors with dimensions \( C \times H \times W \), where \( C \) denotes the number of channels, and \( H \) and \( W \) represent the height and width of the feature maps, respectively.

 For each spatial location \( (h,k) \), the cosine distance \( CD_{channel} \) is calculated as follows:
\[
CD_{channel} = \sum_{k}^{K}\left(\frac{1}{H^{k}W^{k}}\sum_{h}^{H}\sum_{w}^{W}\left(1 - \frac{{f_{t}^{k}}^{T}f_{s}^{k}}{\|f_{t}^{k}\|_{2} \|f_{s}^{k}\|_{2}}\right)\right)
\]

Where: \( f_{t}^{k} \) and \( f_{s}^{k} \) are 1D feature vectors across channels, \( f_{t}^{k}, f_{s}^{k}  \) \( \in \) \( \mathbb{R}^{D^{k} \times 1} \).

For each channel \( d \), the cosine distance \( CD_{spatial} \) is calculated as follows:
\[
CD_{spatial} = \sum_{k}^{K}\left(\frac{1}{D^{k}}\sum_{d}^{D}\left(1 - \frac{{f_{t}^{k}}^{T}f_{s}^{k}}{\|f_{t}^{k}\|_{2} \|f_{s}^{k}\|_{2}}\right)\right)
\]

 Where: \( f_{t}^{k} \) and \( f_{s}^{k} \) represent the channel-wise 2D feature vectors, \( f_{t}^{k}, f_{s}^{k}  \) \( \in \) \( \mathbb{R}^{W^{k} \times H^{k}} \)

\subsection{KL Divergence (KLD)}

We utilize Kullback–Leibler (KL) divergence for feature matching to identify the distributional differences between student and teacher feature maps. By minimizing the KL-Divergence, the student network learns to align its feature distribution with that of the teacher’s feature distribution. KLD captures the non-linear relationship between distributions resulting in a better replication of features across different categories. To address the complexity of multi-class knowledge distillation arising due to distributions across various classes, we implemented channel KL-divergence by taking one-dimensional vectors along the channel dimension. Additionally, the student network must learn the local and global context, to effectively capture the spatial distribution of the feature maps. By utilizing KLD along the spatial dimension we intend to measure the relative entropy between student and teacher spatial feature distribution.

 Let \( f_{t}^{k} \) and \( f_{s}^{k} \) represent the 1D feature maps across channel, \( f_{t}^{k}, f_{s}^{k}  \) \( \in \) \( \mathbb{R}^{D^{k} \times 1 }\) for the teacher and student, respectively. Here, \( \phi \) represents the softmax function which converts the input vector into a probability distribution. 
 
For each spatial location (h, k), \(KLD_{\text{channel}}\) is calculated as follows:

\[
KLD_{\text{channel}} = \sum_{k=1}^{K}\sum_{w=1}^{W}\sum_{h=1}^{H} \phi(f_{t}^k) \cdot \log\left(\frac{\phi(f_{t}^k)}{\phi(f_{s}^k)}\right)
\] 

\[\text{where, }
\phi(f_{t}^k) = \frac{exp(f_{t}^{k})}{\sum_{d=1}^{D}exp(f_{t}^{k})_{d}}
\]

 Let \( f_{t}^{k} \) and \( f_{s}^{k} \) represent channel-wise the 2D feature vectors, \( f_{t}^{k}, f_{s}^{k}  \) \( \in \) \( \mathbb{R}^{W^{k} \times H^{k} }\). 
 
 For each channel d, \(KLD_{\text{spatial}}\) is calculated as follows:

\[
KLD_{\text{spatial}} = \sum_{k=1}^{K}\sum_{d=1}^{D} \phi(f_{t}^k) \cdot \log\left(\frac{\phi(f_{t}^k)}{\phi(f_{s}^k)}\right)
\]
\[\text{where, }
\phi(f_{t}^k) = \frac{exp(f_{t}^{k})}{\sum_{h=1}^{H}\sum_{w=1}^{W}exp(f_{t}^{k})_{w \times h}}
\]

\subsection{Inference Phase}

\begin{figure}[htbp]
  \centering
  \includegraphics[width=1.05\textwidth]{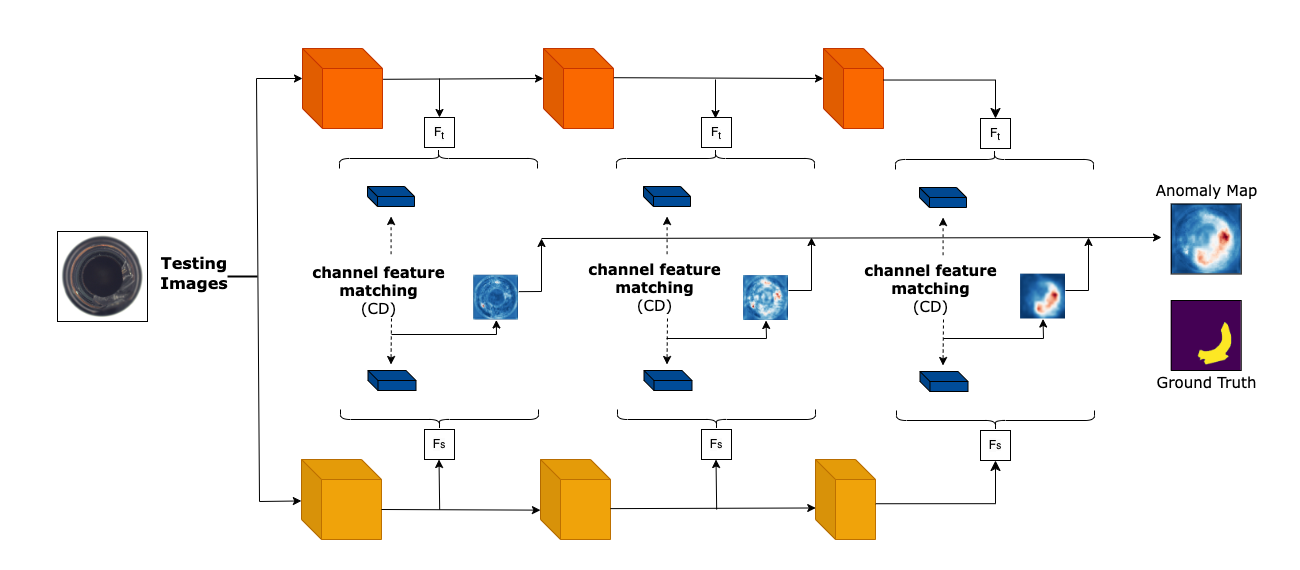}
  \caption{Overview of our teacher-student framework (Inference phase). The orange and yellow blocks represent the $2^{nd}$, $3^{rd}$, and $4^{th}$ convolutional blocks of the teacher and student network respectively. During the inference phase, the anomaly map is created by aggregating upsampled loss maps of each block calculated using cosine distance between the teacher and student feature maps. The progressive formation of the anomaly map for a sample test image (category: bottle) is shown alongside the ground truth.}
  \label{fig:image_label}
\end{figure}
In the training process with anomaly-free (or normal) samples, the student and teacher feature maps come closer to each other in feature space. During inference, we compute the cosine distance between the student and teacher attention learned feature maps. When shown an anomalous image, we get a higher cosine distance between teacher and student as only normal samples are used in training. Since the student network has already undergone attention-based learning of the feature maps during training, spatial and channel attention mechanisms are not utilized in the inference phase. Consequently, the latency of our method remains unchanged over the baseline method \cite{STPFM}. This is a unique feature of our methodology which improves the localisation performance significantly with the same inference time. The detailed analysis of latency for the proposed method is described in the ablation study section.
\vspace{-10 pt}
\section{Experiments and Results}
\subsection{Dataset}
 We have used the MVTec AD dataset \cite{mvtec_dataset} for experimentation purposes. MVtec AD \cite{mvtec_dataset} is a benchmark dataset that contains over 5000 images of various objects and textures such as carpet, leather, etc. used in both image level and pixel level anomaly detection. For training our model, we have used anomaly-free images for each of the 15 categories, whereas for testing purposes both anomaly-free and anomalous images were used. We evaluated our methodology using the metrics: AUC-ROC (Area Under the Receiver Operating Characteristic curve) and PRO (per region overlap). For latency, we calculated the processing time of the test function to generate loss maps for images across individual classes and then computed a weighted average across all classes. 
 
 For the baseline, we used the 15 class MVTec AD \cite{mvtec_dataset} data in the STFPM (Student-Teacher Feature Pyramid Matching). The train set contains 3629 anomaly-free images and the test set consists of 1725 mixed types of images. \hfill\break

\subsection{Implementation}
 For all the experiments, we use a teacher-student architecture, where both the teacher and student networks are based on ResNet-18. The teacher network is pretrained on ImageNet \cite{imagenet}, while the student network is initialized with random weights. We choose the first three convolution blocks of the ResNet-18 architecture, namely conv2\_x, conv3\_x, and conv4\_x, for the knowledge distillation process. All images in our experiments are resized to \(256 \times 256\) and normalized by the mean and variance of ImageNet \cite{imagenet}.We train the network using stochastic gradient descent (SGD) with a learning rate of 0.1 for 400 epochs and a batch size of 32. For hyper-parameters, we simply set \( \lambda = 0.5 \) for KLD. The experiments are implemented using PyTorch on a single GPU node with a Tesla V100-SXM2 16GB GPU card, and the during inference phase latency is measured on the system Macbook Air 2017 (1.8 GHz Dual-Core Intel Core i5) \hfill\break
 \vspace{-20 pt}
\subsection{Training and Testing}
In the training process, we first reshape and transform each training image in the dataset, followed by splitting the good images into training and validation sets by an 80-20 split. After splitting the dataset, we feed the data to the teacher and student models. The teacher model is pre-trained on ImageNet \cite{imagenet}, whereas we have added a Distributed Convolutional Attention Module (DCAM) to the student model after the 2nd, 3rd, and 4th convolution blocks. In each iteration, we compute the loss between the student and teacher feature map channel-wise and spatially. After each epoch, we save the weights that have the minimum validation loss.

 In the testing process, we construct an anomaly map $\Omega^{w \times h}$. We feed the testing image \textbf{I}, $I \in \mathbb{R}^{w\times h \times c}$. Let $f_{t}^{k}$ and $f_{s}^{k}$ be the $k^{th}$ feature map generated by the teacher and student model, respectively. We compute a loss map $\Omega^{k}$ by computing the cosine distance between the student and teacher feature map, which is then upsampled to size $w \times h$ using bilinear interpolation. Final anomaly map $\Omega$ is the element-wise addition of each upsampled loss maps.
\[
\Omega(I) = \sum_{k}^{K} Upsample(\Omega^{k}(I))
\]
 \vspace{-20 pt}
\subsection{Results}
Our study compares the performance of different attention mechanisms and feature-matching metrics in multi-class anomaly detection. Combining the best-performing attention module and feature-matching metrics led to the highest performance, as demonstrated in Table~\ref{table:4}, with AUC-ROC of 95.20\%, PRO of 89.81\%, and an inference time of 0.3169 s per image. Comparing our methodology with the original STFPM \cite{STPFM} approach, as outlined in Table~\ref{table:5}, reveals significant improvements. Our approach surpassed the baseline by 3.92\% in AUC-ROC and 6.8\% in PRO, while maintaining a comparable latency.
\vspace{-4em}
\vspace{-10 pt}
\FloatBarrier
\begin{table*}
\centering
\renewcommand{\arraystretch}{1.4} 
\begin{tabular}{|>{\centering\arraybackslash}m{2.8cm}|*{6}{>{\centering\arraybackslash}m{1.5cm}|}} 
\hline
           & \multicolumn{3}{c|}{\textbf{STFPM}} & \multicolumn{3}{c|}{\textbf{OURS}} \\
\hline
           \textbf{CATEGORY} & \textbf{AUC ROC} & \textbf{PRO} & \textbf{Latency} & \textbf{AUC ROC} & \textbf{PRO} & \textbf{Latency} \\
\hline
\textbf{BOTTLE}     & 0.9423 & 0.8422 & 0.3179 & 0.9681 & 0.9113 & 0.3208 \\
\textbf{CABLE}      & 0.8501 & 0.7154 & 0.317  & 0.8481 & 0.8459 & 0.3122 \\
\textbf{CAPSULE}    & 0.9175 & 0.7122 & 0.3201 & 0.976 & 0.8819 & 0.3109 \\
\textbf{CARPET}     & 0.9794 & 0.9332 & 0.3222 & 0.9887 & 0.9586 & 0.3466 \\
\textbf{GRID}       & 0.9618 & 0.8961 & 0.3165 & 0.975 & 0.9248 & 0.3106 \\
\textbf{HAZELNUT}   & 0.9616 & 0.9128 & 0.3179 & 0.9836 & 0.9415 & 0.3334 \\
\textbf{LEATHER}    & 0.9913 & 0.977  & 0.3375 & 0.9912 & 0.9756 & 0.3111 \\
\textbf{METAL NUT}  & 0.8596 & 0.782  & 0.3329 & 0.9305 & 0.8792 & 0.3123 \\
\textbf{PILL}       & 0.8172 & 0.7476 & 0.3179 & 0.9665 & 0.9133 & 0.3129 \\
\textbf{SCREW}      & 0.9174 & 0.7966 & 0.3154 & 0.9656  & 0.8764 & 0.3139 \\
\textbf{TILE}       & 0.9479 & 0.8839 & 0.3165 & 0.9565 & 0.8701 & 0.3145 \\
\textbf{TOOTHBRUSH} & 0.9343 & 0.7523 & 0.3172 & 0.9789   & 0.8046 & 0.3127 \\
\textbf{TRANSISTOR} & 0.7693 & 0.7869 & 0.3181 & 0.8396 & 0.8815 & 0.3118 \\
\textbf{WOOD}       & 0.9305 & 0.8672 & 0.3152 & 0.9405 & 0.8915 & 0.3159 \\
\textbf{ZIPPER}     & 0.9119 & 0.8468 & 0.3145 & 0.969 & 0.9158 & 0.3148 \\
\hline
\textbf{MEAN}       & 0.9128 & 0.8301 & 0.3198 & \textbf{0.9520} & \textbf{0.8981} & \textbf{0.3169} \\
\hline
\end{tabular}
\vspace{2.5pt}
\caption{Comparison of AUC-ROC, PRO, and Latency (in sec)}
\label{table:5}

\vspace{10pt} 
\vspace{-2em}
\begin{tabular}{|>{\centering\arraybackslash}m{1.9cm}|*{7}{>{\centering\arraybackslash}m{1.5cm}|}} 
\hline
           & \multicolumn{2}{c|}{\textbf{Non KD Based}} & \multicolumn{5}{c|}{\textbf{KD Based}} \\
\hline
            \textbf{Method}
            & \textbf{UniAD} & \textbf{OmniAL} & \textbf{US} & \textbf{STFPM} & \textbf{SNL} & \textbf{RD}& \textbf{Ours} \\

\hline
\textbf{AUC ROC} & 96.8 & 98.3 & 81.8 & 91.28 & 98.7 & 95.0 & \textbf{95.2} \\
\hline
\end{tabular}
\vspace{2.5pt}
\caption{Comparison of AUC-ROC of our approach with some KD Based and Non-KD Based methods}
\label{table:6}
\end{table*}
\FloatBarrier
\vspace{-15 pt}
Although the goal of our study is mainly focused on Knowledge-Distillation based methods, we mentioned some recent multi-class works whose performance may be higher than our approach but they are architecture-heavy and latency-intensive (discussed in Section 2). Comparing the Knowledge-Distillation (KD) based methods, SNL shows a higher performance than ours again with a latency trade-off as it involved computation-intensive operations (Section 2) and a WideResNet-50 as backbone while our approach is built on the baseline STFPM with an unchanged backbone of ResNet-18. RD also has a close performance to ours with the same disadvantage because of a WideResNet-50 architecture with the presence of a bottleneck to project the teacher model's high dimensional representation into a low-dimensional space, adding even more to the latency.

\subsection{Ablation Study}
In this section, we present the results of ablation studies conducted to evaluate the impact of different components in our approach. We systematically analyze the performance of our model by selectively removing and altering specific components, including Distributed Convolutional Attention Module (DCAM) and loss metrics. Through these experiments, we aim to gain insights into the individual contributions of each component.

\subsubsection{DCAM Evaluation}

We first evaluated the DCAM module by conducting three sets of experiments: (1) using only channel attention, (2) using only spatial attention, and (3) combining both channel and spatial attention. All of the above experiments involved only MSE as loss.

\FloatBarrier
\begin{table}[h]
\centering

\setlength{\tabcolsep}{3.5pt} 
\renewcommand{\arraystretch}{1.2} 
\begin{tabular}{|>{\centering\arraybackslash}m{3.2cm}|>{\centering\arraybackslash}m{1.8cm}|>{\centering\arraybackslash}m{1.8cm}|>{\centering\arraybackslash}m{1.8cm}|} 
\hline
 \textbf{DCAM}
 & \textbf{AUC-ROC} & \textbf{PRO}& \textbf{Latency} \\
\hline
Channel   & \textbf{0.9412}   & 0.8835 & \textbf{0.3181}\\
Spatial   & 0.9336   & 0.8629 & 0.3182\\
Combined (Channel + Spatial)  & 0.9367   & \textbf{0.8871} & 0.3187 \\
\hline
\end{tabular}
\vspace{2.5pt}
\caption{DCAM ablation study}
\label{table:1}
\end{table}
\FloatBarrier

As shown in Table~\ref{table:1}, the integration of the channel attention module gave better results compared to spatial and combined attention modules, achieving an AUC-ROC of 94.12\% and PRO of 88.35\%.  \hfill \break

\vspace{-5pt}

\subsubsection{Feature Matching Analysis}
Next, we compared Cosine Distance (CD) and Kullback-Leibler Divergence (KLD) for both channel and spatial feature matching. The AUC-ROC, PRO and latency (in sec) results for each method are presented in Tables \ref{table:2} and \ref{table:3}, respectively.

\begin{table}[h]
\centering
\setlength{\tabcolsep}{3.5pt} 
\renewcommand{\arraystretch}{1.2} 
\begin{tabular}{|>{\centering\arraybackslash}m{1.8cm}|>{\centering\arraybackslash}m{1.8cm}|>{\centering\arraybackslash}m{1.8cm}|>{\centering\arraybackslash}m{1.8cm}|} 
\hline
\textbf{CD}
& \textbf{AUC-ROC} & \textbf{PRO}& \textbf{Latency} \\
\hline
Channel  & \textbf{0.9392}  &\textbf{0.8796} & \textbf{0.3180}   \\
Spatial  & 0.8849  & 0.8081 & 0.3180 \\
\hline
\end{tabular}
\vspace{1pt}
\caption{CD based Feature Matching along channel and spatial dimension}
\label{table:2}
\end{table}

\vspace{-2em}
\begin{table}[h]
\centering
\setlength{\tabcolsep}{3.5pt} 
\renewcommand{\arraystretch}{1.2} 
\begin{tabular}{|>{\centering\arraybackslash}m{1.8cm}|>{\centering\arraybackslash}m{1.8cm}|>{\centering\arraybackslash}m{1.8cm}|>{\centering\arraybackslash}m{1.8cm}|} 
\hline
\textbf{KLD}
& \textbf{AUC-ROC} & \textbf{PRO}& \textbf{Latency} \\
\hline
Channel  & 0.938   & 0.8845 & 0.3181\\
Spatial  & \textbf{0.9467}  & \textbf{0.886}& \textbf{0.3180} \\
\hline
\end{tabular}
\vspace{1pt}
\caption{KLD based feature matching along channel and spatial dimensions}
\label{table:3}
\end{table}

The results of Cosine Distance (CD) (Table~\ref{table:2}) and KL-Divergence (KLD) (Table~\ref{table:3}) revealed that cosine distance is a better feature-matching metric in the channel dimension, whereas KL-Divergence is more effective in the spatial dimension.

\subsubsection{Integration of Combined Methods}
Finally, based on the results from the previous experiments, we picked the best performing methodologies and designed another set of experiments combining CD and KLD for feature matching with and without the DCAM. The AUC-ROC, PRO and Latency results, presented in Table ~\ref{table:4}, demonstrate the effectiveness of our approach with and without the inclusion of channel attention module.
\FloatBarrier
\begin{table}[h] 
\centering
\setlength{\tabcolsep}{3.5pt} 
\renewcommand{\arraystretch}{1.3} 
\begin{tabular}{|>{\centering\arraybackslash}m{1.8cm}|>{\centering\arraybackslash}m{4.5cm}|>{\centering\arraybackslash}m{1.8cm}|>{\centering\arraybackslash}m{1.4cm}|>{\centering\arraybackslash}m{1.4cm}|} 
\hline
\textbf{DCAM (channel)} & \textbf{LOSSES} & \textbf{AUC-ROC} & \textbf{PRO}& \textbf{Latency} \\
\hline
\ding{51} & Channel(CD) + Spatial(KLD) & \textbf{0.9520} & \textbf{0.8981} & \textbf{0.3169} \\
\ding{55} & Channel(CD) + Spatial(KLD) & 0.9514 & 0.8859 &  0.3172\\
\hline
\end{tabular}
\vspace{2.5pt}
\caption{Combined results of Channel Attention and best of CD and KLD}
\label{table:4}
\end{table}
\FloatBarrier

Here, we conclude that our combination of Channel-wise DCAM with Channel-wise CD and Spatial KD showed the highest performance achieving an AUROC of 95.20\% with a latency of 0.317 secs.

\subsection{Latency Comparison Analysis}
We observe consistent latency across all proposed methodologies, as depicted in Table ~\ref{table:1}~\ref{table:2}~\ref{table:3}~\ref{table:4}~\ref{table:5}. This uniformity persists due to the exclusion of the Distributed Convolutional Attention Module (DCAM) during the inference phase. We solely compute the cosine distance between feature maps, maintaining a comparable model complexity to that of the (STFPM) method \cite{STPFM}. Moreover, leveraging ResNet-18 as the backbone ensures uniformity in learnable parameters. 

\section{Conclusion}
We present an attention-based feature-matching technique and incorporate it into the student-teacher anomaly detection architecture. Given a powerful network pre-trained on image classification as the teacher, we use its different levels of features to guide a student network, introducing the concept of important features to the network, and enabling the student to prioritize learning crucial features. This ensures that the student network effectively learns the distribution of anomaly-free images. In multi-class scenario, the normal distribution across multiple categories become more complex than that in one-class scenarios. So the distillation needs more constraints for better learning of student features which is accomplished by learning of convolutional attention masks over the feature representations. Our proposed solution is not only more efficient and scalable, as we utilize only one model for all classes unlike other approaches, but also demonstrates comparable latency.
Through hierarchical feature matching, our approach demonstrates the capability to detect anomalies of varying sizes with a single forward pass. Experimental evaluation conducted on the MVTec AD dataset validates the superiority of our method over the state-of-the-art alternatives.

\section{Acknowledgement}
This work is supported by Hitachi India Pvt. Ltd.

%
%
%
\bibliographystyle{splncs04}
\bibliography{ref}

\begin{thebibliography}{10}
\providecommand{\url}[1]{\texttt{#1}}
\providecommand{\urlprefix}{URL }
\providecommand{\doi}[1]{https://doi.org/#1}

\bibitem{IUDS}
Bergmann, P., Löwe, S., Fauser, M., Sattlegger, D., Steger, C.: Improving unsupervised defect segmentation by applying structural similarity to autoencoders. In: Proceedings of the 14th International Joint Conference on Computer Vision, Imaging and Computer Graphics Theory and Applications. SCITEPRESS - Science and Technology Publications (2019). \doi{10.5220/0007364503720380}, \url{http://dx.doi.org/10.5220/0007364503720380}

\bibitem{DLBDD}
Jezek, S., Jonak, M., Burget, R., Dvorak, P., Skotak, M.: Deep learning-based defect detection of metal parts: evaluating current methods in complex conditions. In: 2021 13th International Congress on Ultra Modern Telecommunications and Control Systems and Workshops (ICUMT). pp. 66--71 (2021). \doi{10.1109/ICUMT54235.2021.9631567}

\bibitem{DLADACM}
Pang, G., Shen, C., Cao, L., Hengel, A.V.D.: Deep learning for anomaly detection: A review. ACM Computing Surveys  \textbf{54}(2),  1–38 (Mar 2021). \doi{10.1145/3439950}, \url{http://dx.doi.org/10.1145/3439950}

\bibitem{ADACM}
Nadipuram R.~Prasad, Salvador Almanza-Garcia, T.T.L.: Anomaly detection. Computers, Materials \& Continua  \textbf{14}(1),  1--22 (2009). \doi{10.3970/cmc.2009.014.001}, \url{http://www.techscience.com/cmc/v14n1/22504}

\bibitem{MDDM}
Saberironaghi, A., Ren, J., El-Gindy, M.: Defect detection methods for industrial products using deep learning techniques: A review. Algorithms  \textbf{16}(2) (2023). \doi{10.3390/a16020095}, \url{https://www.mdpi.com/1999-4893/16/2/95}

\bibitem{12}
Zhou, C., Paffenroth, R.C.: Anomaly detection with robust deep autoencoders. In: Proceedings of the 23rd ACM SIGKDD international conference on knowledge discovery and data mining. pp. 665--674 (2017)

\bibitem{13}
Defard, T., Setkov, A., Loesch, A., Audigier, R.: Padim: a patch distribution modeling framework for anomaly detection and localization. In: International Conference on Pattern Recognition. pp. 475--489. Springer (2021)

\bibitem{14}
Li, C.L., Sohn, K., Yoon, J., Pfister, T.: Cutpaste: Self-supervised learning for anomaly detection and localization. In: Proceedings of the IEEE/CVF conference on computer vision and pattern recognition. pp. 9664--9674 (2021)

\bibitem{15}
Zavrtanik, V., Kristan, M., Sko{\v{c}}aj, D.: Draem-a discriminatively trained reconstruction embedding for surface anomaly detection. In: Proceedings of the IEEE/CVF International Conference on Computer Vision. pp. 8330--8339 (2021)

\bibitem{16}
Zhao, Y.: Just noticeable learning for unsupervised anomaly localization and detection. In: 2022 IEEE International Conference on Multimedia and Expo (ICME). pp. 01--06. IEEE (2022)

\bibitem{17}
You, Z., Cui, L., Shen, Y., Yang, K., Lu, X., Zheng, Y., Le, X.: A unified model for multi-class anomaly detection. Advances in Neural Information Processing Systems  \textbf{35},  4571--4584 (2022)

\bibitem{18}
Zhao, Y.: Omnial: A unified cnn framework for unsupervised anomaly localization. In: Proceedings of the IEEE/CVF Conference on Computer Vision and Pattern Recognition. pp. 3924--3933 (2023)

\bibitem{19}
Bergmann, P., Fauser, M., Sattlegger, D., Steger, C.: Uninformed students: Student-teacher anomaly detection with discriminative latent embeddings. In: Proceedings of the IEEE/CVF conference on computer vision and pattern recognition. pp. 4183--4192 (2020)

\bibitem{20}
Deng, H., Li, X.: Anomaly detection via reverse distillation from one-class embedding. In: Proceedings of the IEEE/CVF Conference on Computer Vision and Pattern Recognition. pp. 9737--9746 (2022)

\bibitem{STPFM}
Wang, G., Han, S., Ding, E., Huang, D.: Student-teacher feature pyramid matching for anomaly detection. arXiv preprint arXiv:2103.04257  (2021)

\bibitem{hinton2015distilling}
Hinton, G., Vinyals, O., Dean, J.: Distilling the knowledge in a neural network. arXiv preprint arXiv:1503.02531  (2015)

\bibitem{imagenet}
Deng, J., Dong, W., Socher, R., Li, L.J., Li, K., Fei-Fei, L.: Imagenet: A large-scale hierarchical image database. In: 2009 IEEE conference on computer vision and pattern recognition. pp. 248--255. Ieee (2009)

\bibitem{mvtec_dataset}
Bergmann, P., Batzner, K., Fauser, M., Sattlegger, D., Steger, C.: The mvtec anomaly detection dataset: a comprehensive real-world dataset for unsupervised anomaly detection. International Journal of Computer Vision  \textbf{129}(4),  1038--1059 (2021)

\bibitem{deng2024structural}
Deng, H., Li, X.: Structural teacher-student normality learning for multi-class anomaly detection and localization (2024)

\bibitem{xu2020knowledge}
Xu, G., Liu, Z., Li, X., Loy, C.C.: Knowledge distillation meets self-supervision. In: European conference on computer vision. pp. 588--604. Springer (2020)

\bibitem{woo2018cbam}
Woo, S., Park, J., Lee, J.Y., Kweon, I.S.: Cbam: Convolutional block attention module. In: Proceedings of the European conference on computer vision (ECCV). pp. 3--19 (2018)

\bibitem{article}
Dubey, V., Saxena, A.: A cosine-similarity mutual-information approach for feature selection on high dimensional datasets. Journal of Information Technology Research  \textbf{10},  15--28 (01 2017). \doi{10.4018/JITR.2017010102}

\end{thebibliography}

\end{document}